\documentclass[%
aip,
rsi,%
amsmath,amssymb,
reprint,%
]{revtex4-1}

\usepackage{graphicx}% Include figure files
\usepackage{dcolumn}% Align table columns on decimal point
\usepackage{bm}% bold math
\usepackage{algorithm}%
\usepackage{algorithmicx}%
\usepackage{algpseudocode}%
\usepackage{float}
\usepackage[colorlinks=true, linkcolor=blue, citecolor=blue, urlcolor=blue]{hyperref}

\begin{document}
	
	\title{Impact of white noise in artificial neural networks trained for classification: performance and noise mitigation strategies}
	
	\author{Nadezhda  Semenova}
	\email{semenovani@sgu.ru}
	\affiliation{Saratov State University, Astrakhanskaya str. 83, Saratov 410012, Russia}%Lines break automatically or can be forced with \\

	\author{Daniel Brunner}
	\affiliation{FEMTO-ST Institute/Optics Department, CNRS \& University Franche-Comt\'e, \\15B avenue des Montboucons,
		Besan\c con Cedex, 25030, France%\\This line break forced% with \\
	}%

	\date{\today}% It is always \today, today,
	%  but any date may be explicitly specified
	
	\begin{abstract}
		
In recent years, the hardware implementation of neural networks, leveraging physical coupling and analog neurons has substantially increased in relevance. Such nonlinear and complex physical networks provide significant advantages in speed and energy efficiency, but are potentially susceptible to internal noise when compared to digital emulations of such networks. In this work, we consider how additive and multiplicative Gaussian white noise on the neuronal level can affect the accuracy of the network when applied for specific tasks and including a softmax function in the readout layer. We adapt several noise reduction techniques to the essential setting of classification tasks, which represent a large fraction of neural network computing. We find that these adjusted concepts are highly effective in mitigating the detrimental impact of noise.	
		
	\end{abstract}
	
	\maketitle
	
\section{Introduction}\label{sec:intro}
Artificial neural networks (ANNs) play an important role in various fields such as pattern recognition, data analysis, industrial control and complex problem solving. They can be trained on a large amount of data and take reasonably balanced decision based on this information. With these capabilities, they are successfully used in medicine \cite{Amato2013, Salahuddin2022, Sarvamangala2022, Celard2023}, finance \cite{Lazcano2023, Li2023}, technology \cite{Aghbashlo2015, Almonacid2017, Yang2023}, biology \cite{Marabini1994, Suzuki2011, Samborska2014}, and for predicting the behavior of complex, chaotic systems \cite{Wang2024}. The use of ANNs makes it possible to accelerate and improve decision-making processes, which significantly increases the efficiency and accuracy of systems.

However, substantial challenges with energy efficiency, speed and scalability of ANNs are some of the main limitations of their application in various fields \cite{Markovic2020}. For example, complex tasks such as large-volume or real-time data processing require significant computing resources and hence energy. In addition, augmenting ANNs to find complex patterns can lead to problems with scalability and computing resource management \cite{Christensen2022}.

Motivated by these limitations, there is a growing amount of research into developing more energy-efficient ANN architectures, optimizing computation and developing specialized hardware to perform ANN operations, called hardware neural networks \cite{Seiffert2004, Misra2010, Bouvier2019}. These efforts aim at improving the efficiency and performance of ANNs and open the prospect of their wider application in various fields. In hardware ANNs, also called in memory computing, the artificial neurons and connections between them are based on physical principles such as optical \cite{Wang2022, Ma2023}, memristive \cite{Tuma2016,Lin2018,Xia2019}, spin-torque \cite{Torrejon2017}, Mach–Zehnder interferometer \cite{Shen2017, Cem2023}, photoelectronic \cite{Chen2023} or coherent silicon photonics \cite{MourgiasAlexandris2022} effects.

A major consequence of such networks in comparison to digital network emulations is that analogue hardware is always prone to a certain level of noise. Noise in hardware ANNs can arise from a variety of sources. Regardless of its origin, such noise can have a negative impact on the accuracy of the ANN, as it can cause errors in the transmission and transformation of information. There is a substantial amount of literature on mitigation strategies for noise in the input signal of ANNs \cite{Maas2012, Burger2012, Seltzer2013, Yue2022}, but in the case of hardware ANNs the general context changes as here noise arises internally of the network. There are several papers that describe the properties of various internal noise sources in hardware ANNs \cite{Dolenko1993, Dibazar2006, Soriano2015, Janke2020, Nurlybayeva2022, Ma2023}. 

In our previous papers \cite{Semenova2019, Semenova2022NN}, we have studied the impact of white Gaussian internal noise on simplified and trained neural networks with linear \cite{Semenova2019} and nonlinear \cite{Semenova2022NN} activation functions. Moreover, we developed an analytical description for predicting the noise level in the output of ANN with internal noise. Further, in Ref. \cite{Semenova2022Chaos}, we proposed several techniques how to reduce different types of internal noise on untrained networks. However, this does not include the essential step of generalizing these techniques towards application scenarios. This essential development we report here, and focus on the hardware ANN's performance in the context of classifications tasks, with the commonly applied softmax function in the final layer. Again, we consider a wide range of different noise types, yet the impact of thresholding for classification, as well as the probabilistic transformation through the softmax demand substantial innovation with regards to analysis as well as noise mitigation techniques.

This paper starts explaining the considered trained deep ANN (Sect.~\ref{sec:system_ANN}). We consider the impact of several cases of white Gaussian noise in the deep hardware ANN trained for digit recognition, and we evaluate the noise impact in terms of accuracy degradation (Sect.~\ref{sec:noise_impact}). Noise types are additive and multiplicative, correlated and uncorrelated (they are explained in details in Sect.~\ref{sec:system_noise}). We then apply two noise mitigation techniques to different cases of internal noise and explain how these techniques can be realized schematically and in terms of connection matrices for already trained network (Sect.~\ref{sec:pooling}, \ref{sec:ghost}). In Supplementary materials we suggest how the connection matrices can be modified according to these techniques using Python code as an example.

The main difference from our previous work is that here we consider different noise intensities rather than one set, which allows us to increase the range of application of the results to different types of hardware ANNs. In addition, we previously considered the effect of noise from the point of view of only the signal-to-noise ratio (SNR) of the output signal. In the case of classifying ANNs, this is not entirely correct, since the last layer of such networks often uses the softmax function, for which it is not the output signal itself that is important, but the sequence number of the neuron with the maximum output signal. In this article, we consider the impact of various noises of different intensities on the accuracy of the classification network. Therefore, the conclusions proposed in this article are fundamentally new comparing to all our previous works.

\section{system under study}\label{sec:system}
\subsection{Trained ANN}\label{sec:system_ANN}
We study the impact of internal noise on a trained hardware ANN and show how our noise mitigation techniques can be applied there. For this purpose, we trained a deep ANN with feedforward signal propagation. For training, we took the frequently used handwritten digit recognition task using the MNIST database as an representative example \cite{LeCunSite}. This database contains the images of size $28\times 28$ pixels in a gray scale. Thus, the input layer of our network contains 784 neurons with real input value in range from 0 to 1, and our ANNs has 10 output neurons in the final layer to identify the digit's number. Finally, the output neurons exhibit the softmax normalization, for which it is not the output signal itself that is important, but the sequence number of the neuron with the maximum output signal. As we will show, this nonlinear transformation substantially influences the impact of noise onto the system's output and the final hardware ANN's computational result.

\begin{figure}[h]
	\includegraphics[width=\linewidth]{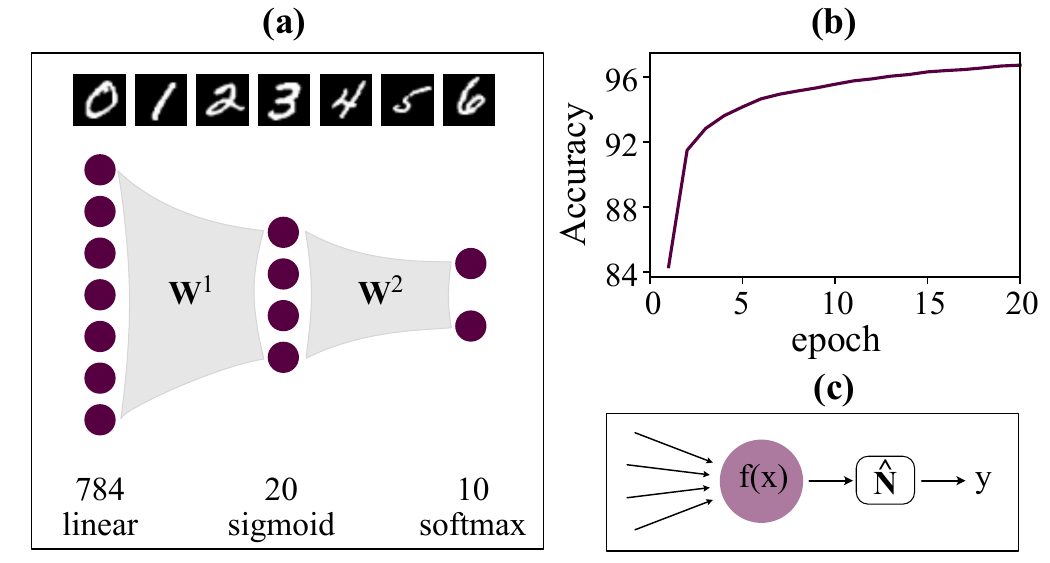}
	\caption{\label{fig:scheme} Schematic representation of considered ANN (a) and process of its training (b). Panel (c) shows how the internal noise is included into one neuron.}
\end{figure}

In this article, we develop and adjust noise mitigation techniques for hardware ANNs trained for classification, which is among the most relevant application scenarios. Therefore, for this proof of concept demonstration, we consider a simple deep network with a single hidden layer comprising 20 neurons with a sigmoid activation function. However, our approaches can be extended to more complex and deeper systems, or to a layer-wise architecture where hybrid ANN concepts are employed. This network is schematically shown in Fig.~\ref{fig:scheme}(a). The network was trained using Tensorflow/Keras library employing an ``adam'' optimizer and ``categorial-crossentropy'' as the loss function. The convergence of training during 20 epoches with a 5\% validation split is given in Fig.~\ref{fig:scheme}(b). The final accuracy tends to 97.01\% on training dataset (60,000 images) and 95.17\% on testing dataset (10,000 images).

\subsection{Internal noise}\label{sec:system_noise}

The method describing a neuron's internal noise is identical to our previous works \cite{Semenova2019, Semenova2022NN, Semenova2022Chaos}. The types of noise, their intensities and methods of introduction were obtained from a hardware implementation of an ANN in an optical experiment, proposed in Ref.~\cite{Bueno2018}. We consider different noise intensities in order to make results more general and applicable to other hardware networks. 

Figure \ref{fig:scheme}(c) schematically illustrates the effect of noise on one neuron and at what stage noise is introduced. A neuron receives the summed input signal from neurons of the previous layer to create its internal state $x$. Then the activation function is applied, and here we use sigmoid function of the type $f(x)=1/(1+e^{-x})$. Usually this is the output signal of the neuron, but in order to include the effect of noisy, analog neurons,  noise operator $\mathbf{\hat{N}}$ is applied to this signal, leading to a final noisy signal of the form $y=\mathbf{\hat{N}} f(x)$. In matrix form this can be described as 
\begin{equation}\label{eq:one_neuron}
	\mathbf{Y}^n = \mathbf{\hat{N}} f(\mathbf{X}^n), \ \ \ \ \mathbf{X}^n = \mathbf{Y}^{n-1} \cdot \mathbf{W}^{n-1} ,
\end{equation}
where $n$ is the layer's number. For example, for hidden layer $n=2$, $\mathbf{X}^n$ is the vector of internal states of neurons on layer $n$, obtained from the values $\mathbf{Y}^{n-1}$ from the previous layer $(n-1)$. The number neurons in hidden layer is equal to $k=20$ for the network suggested in Fig.~\ref{fig:scheme}(a) and the length of vector $\mathbf{X}^n$ is therefore equal to $k=20$. The input signal of the $i$th neuron from the $n$th layer $X^n_i$ is determined in accordance with the outer matrix product illustrated by $\cdot$ between the output signal vector from the previous layer $\mathbf{Y}^{n-1}$ and the coupling matrix $\mathbf{W}^{n-1}$ connecting layers $(n-1)$ and $n$. In Eq.~(\ref{eq:one_neuron}), we describe the effect of noise using noise operator $\mathbf{\hat{N}}$ at the stage where we introduce noise. Next we will look at what this operator is depending on different types of noise. 

In a photonic experimental implementation of a neural network, it was discovered that there are two types of noise in affecting a single neuron: additive noise and multiplicative noise, mathematically described by
\begin{equation}\label{eq:noise}
	\begin{array}{c}
		Y^n_i = f(X^n_i) + \sqrt{2D_A}\xi_A(t,i), \\
		Y^n_i = f(X^n_i) \cdot\big(1+ \sqrt{2D_M}\xi_M(t,i) \big).
	\end{array}
\end{equation}
Thus, additive noise (with indices `A') is added to the noise-free output signal, while the multiplicative noise (with indices `M') is multiplied on it. The notation $\xi$ corresponds to white Gaussian noise with zero mean and variance equal to unity. Its multiplier $\sqrt{2D}$ determines the overall variance equal to $2D$, and $D$ is usually refereed to as the intensity of the noise source. The subscripts `A' and `M' for variables $\xi$ and $D$ correspond to additive and multiplicative noise, respectively. Since $\xi$ has a zero mean, simply multiplying it by the desired signal can lead to losing the entire signal, hence multiplicative noise is introduced by multiplication with $(1+\xi)$.

The previous provides a classification of noise depending on the effect on one isolated neuron. Now let us consider at the types that describe the effect on a group of neurons. As in our previous works \cite{Semenova2019, Semenova2022NN}, we focus on the classification according to uncorrelated and correlated noise within a population of neurons. For both types, the noise values will be different over time $t$. Here, we are not really referring to time, but to different values of input information.
% of digits that are transmitted to the network input. 
The separation of noise types occurs depending on how the noise affects the layer with neurons. If all neurons within one layer receive the same noise value, then we will call such noise correlated and denote it using the superscript `C': $\xi^C(t)$. If all neurons $i$ within one layer receive different noise values, then this is uncorrelated noise $\xi^U(t,i)$. Thus, we consider in total four types of noise: correlated additive $\xi^C_A(t)$ and multiplicative $\xi^C_M(t)$ noise, and uncorrelated additive $\xi^U_A(t,i)$ and multiplicative $\xi^U_M(t,i)$ noise controlling by the noise intensities $D^C_A$, $D^C_M$, $D^U_A$, $D^U_M$, respectively.

\section{Network performance depending on the impact of noise}\label{sec:noise_impact}
\begin{figure}[h]
	\includegraphics[width=\linewidth]{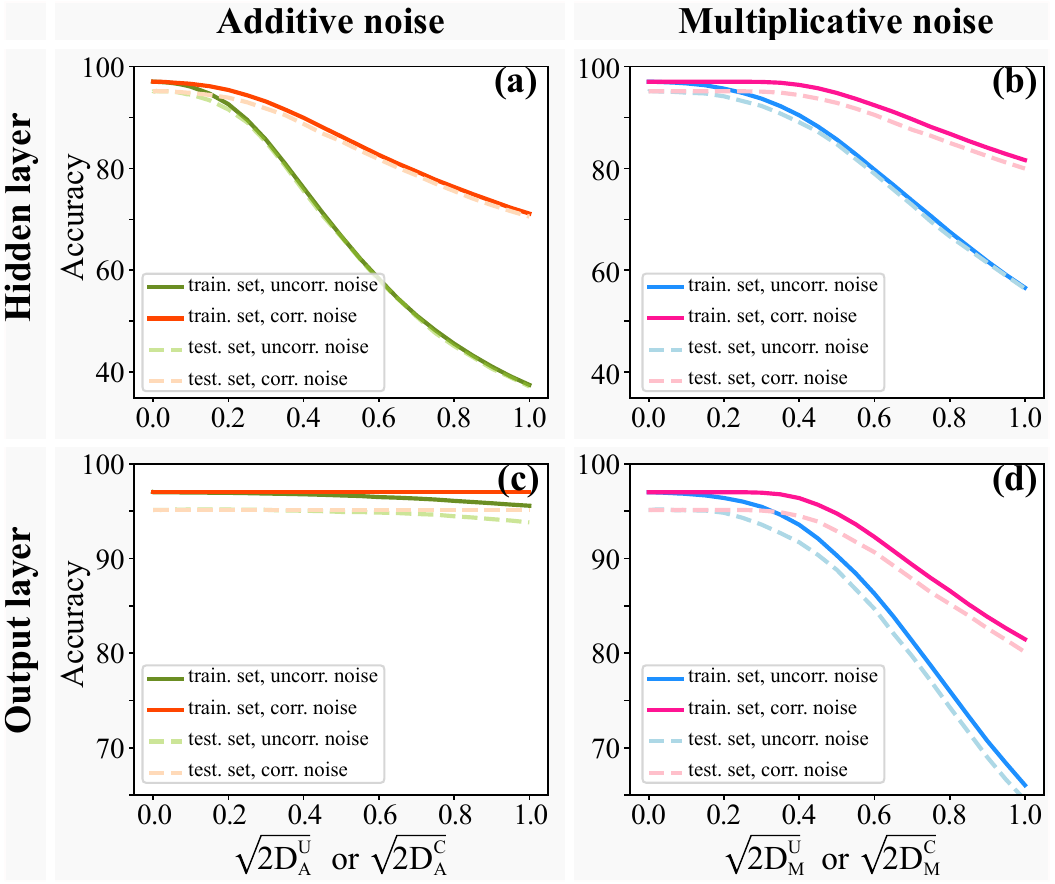}
	\caption{\label{fig:noise_impact} Changes in accuracy of trained ANN depending on noise intensities of four separate noise sources additive uncorrelated noise $D^U_A$ (green in (a,c)), additive correlated noise $D^C_A$ (orange in (a,c)), multiplicative uncorrelated noise $D^U_M$ (blue in (b,d)) and multiplicative correlated noise $D^C_M$ (pink in (b,d)). The noise is introduced into hidden layer in the top panels, and into the last layer in the bottom panels. Solid curves correspond to training dataset, while dashed curves were prepared for testing dataset.}
\end{figure}

Figure \ref{fig:noise_impact} contains training and testing accuracies of the same trained noisy hardware ANN. Here, we consider the impact of noise isolated either to the hidden layer only (top panels) or the in output layer only (bottom panels), while left panels contain the results for additive noise and right panels show the impact of multiplicative noise. All four plots show the impact of different noise intensities on the accuracy of ANN in almost the same scale. 

Figure \ref{fig:noise_impact} shows that networks trained for multiclass classification are quite robust against the impact of weak noise, which was also our finding in \cite{Semenova2022NN}. There is almost no change in accuracy for any type of noise with noise intensity $\sqrt{2D}<0.2$ or $D<0.02$. In this context, it is noteworthy that the largest noise intensity in a non-noise optimized hardware implementation in Ref.\cite{Bueno2018} was around $10^{-3}$. 

As for relatively high noise intensities, the results are as follows. Additive noise (both correlated and uncorrelated) in the output layer has almost no impact on the network performance (see Fig.~\ref{fig:noise_impact}(c)). However, additive noise in hidden layer leads to quite a pronounced decrease in accuracy (Fig.~\ref{fig:noise_impact}(a)). In both, Fig.~\ref{fig:noise_impact} (a) and (c), the dependencies for correlated noise are shown in orange, while green color corresponds to uncorrelated noise. The more pronounced impact in the hidden layer is potentially caused by the particularities of classification tasks in combination with the widely employed softmax function in the output
layer. This is a new and highly relevant finding. In the case of noise in the output layer, all outputs are shifted by the same noise value, and the neuron with maximal output remains the same as without noise. Therefore, noise in hidden layer is the most critical aspect for hardware ANNs applied to classification tasks.

In this classification context, multiplicative noise, however, affects the hidden and the output layer similarly, see Fig.~\ref{fig:noise_impact}(b) and (d), respectively. In both panels, pink data correspond to correlated multiplicative noise, while blue data corresponds to uncorrelated multiplicative noise. At the same time, it should be noted that the accuracy is changed only when $\sqrt{2D}>0.4$ hence for rather strong noise compared to the one usually obtained in hardware ANNs. Again, the exponential normalization by softmax serves as a strongly noise-suppressing function. 

\section{Noise mitigation: Neuron pooling}\label{sec:pooling}
\begin{figure}[h]
	\includegraphics[width=\linewidth]{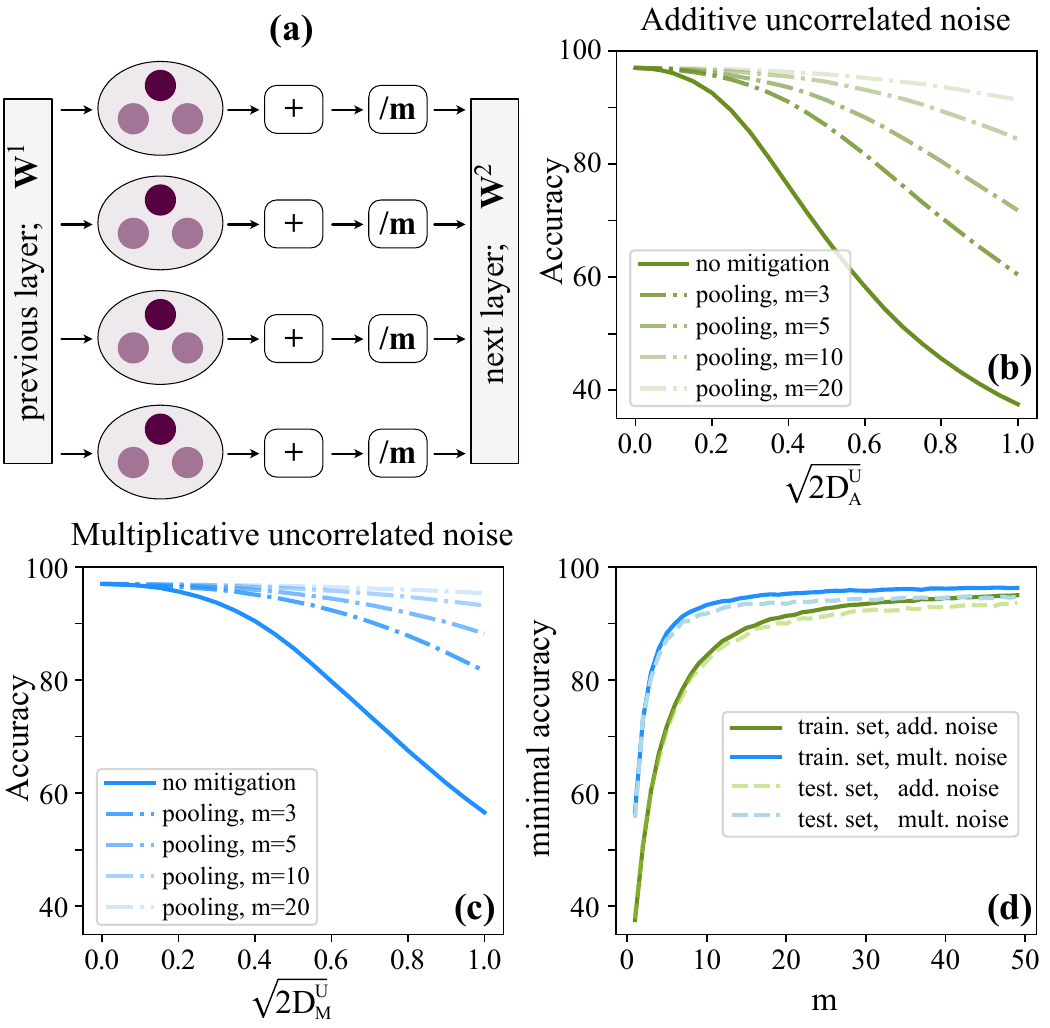}
	\caption{\label{fig:pool} Application of pooling technique to ANN with noisy neurons in hidden layer. Panel (a) shows the scheme of pooling technique with $m=3$. Panels (b) and (c) shows how the dependency of final accuracy on intensity of uncorrelated additive (b) and multiplicative (c) noise. In both panels, solid dark curves correspond to original case without pooling, while the remaining curves correspond to the results of using pooling technique with $m=3$, $5$, $10$ and $20$ (larger $m$ -- lighter color). Panel (d) shows how does the minimum accuracy change depending on $m$ for uncorrelated additive (green) and multiplicative (blue) noise. }
\end{figure}

In our previous work \cite{Semenova2022Chaos} we have suggested several ways how noise can be suppressed in hardware ANNs. The first technique was pooling populations of neurons, an efficient approach to suppress uncorrelated noise. This method consists of combining several neurons into a distinct subgroups called pools. Each unit inside a pool of $m$ neurons receives the same input. The impact of such pooling for trained hardware ANNs is shown in Fig.~\ref{fig:pool}(a) for pooling with $m=3$. 

If the number of noisy neurons in hidden layer is $k$ (here $k=20$), then a corresponding pooled layer comprises after $m\cdot k$ noisy neurons. In order to implement this pooling operations numerically, we repeat the trained connection matrix $m$ and rescale its entries by $1/m$. 
After this operation and for our here leveraged topology, the new sizes of matrices $\mathbf{W}^1$ and $\mathbf{W}^2$ become $(784\times 20m)$ and $(20m\times 10)$, respectively. The description of the way how this can be realized in terms of Python code is given in Supplementary materials. It is also important to underline, that the network is not retrained after this technique. Here we use the connection matrices $\mathbf{W}^1$ and $\mathbf{W}^2$ obtained after training the network and repeat them $m$ times in order to realize pooling.

Figure~\ref{fig:pool} shows the results of application of pooling technique for four different values of $m$ for uncorrelated additive (b) and multiplicative (c) noise in the hidden layer. Here, we show the accuracy only on the training dataset, but for testing dataset the results are similar. In Fig.~\ref{fig:pool}(b,c), the solid lines correspond to the original case without pooling. As can be seen, larger $m$ lead to a significantly reduced sensitivity of the hardware ANN's accuracy for different noise intensities, where pooling 10 neurons makes the system robust even for noise amplitudes reaching $\sqrt{2D}=0.5$. This conclusion is valid for multiplicative as well as additive uncorrelated noise in the context of classification with softmax normalization.

The dependencies of accuracy on noise intensity are nonlinear, and in order to show the improvement of accuracy using the pooling technique, we plotted the dependencies of minimal accuracy on $m$ in Fig.~\ref{fig:pool}(d). As follows from Fig.~\ref{fig:pool}(b,c), these minimal accuracies can be obtained for the largest considered noise intensity $\sqrt{2D^U_A}=1$ or $\sqrt{2D^U_M}=1$. It can be clearly seen from this panel, that for large $m$ even the worst accuracy tends to the initial accuracy level of the noise-free ANN.

The variants of applying the pooling technique in terms of Python code are given in Supplementary materials.

\section{Noise mitigation. Ghost neuron}\label{sec:ghost}
\begin{figure}[h]
	\includegraphics[width=\linewidth]{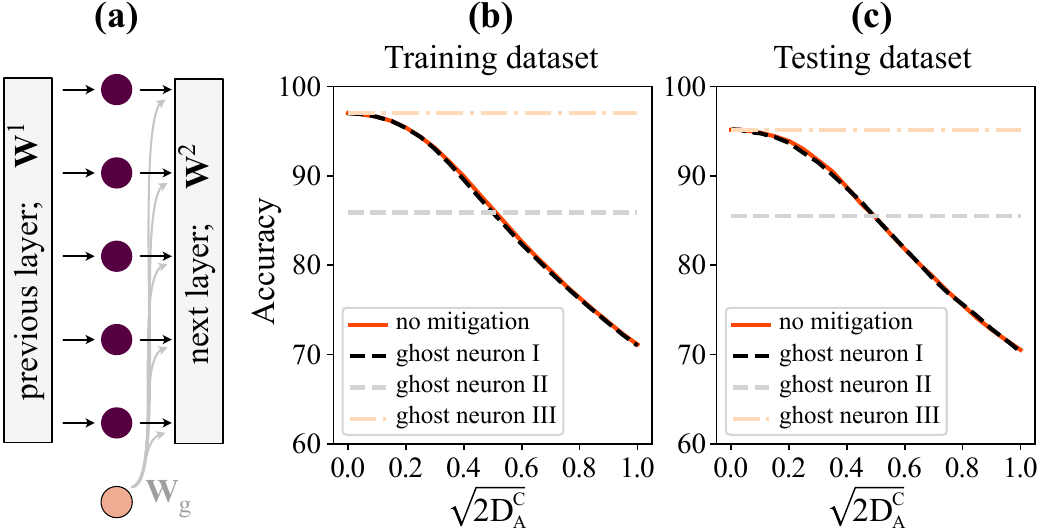}
	\caption{\label{fig:ghost} Application of ghost neuron technique to ANN with noisy neurons in hidden layer. Panel (a) shows the scheme of ghost neuron technique. Panels (b) and (c) shows how the impact of additive correlated noise can be changed using ghost neurons of all three types. Panel (b) and (c) were prepared for training and testing datasets, respectively. }
\end{figure}

The other noise mitigation technique in Ref.~\cite{Semenova2022Chaos} was referred to as ghost neurons, and the concept is schematically illustrated in Fig.~\ref{fig:ghost}. Here, we describe how it can be adapted in the context of trained network and its accuracy. The motivation of this technique is to suppress correlated additive noise by adding one additional neuron into the noisy layer, and to have this neuron not connected with the previous layer. This means that this neuron does not receive any input signal, which led to the nomenclature ``ghost neuron''. Importantly, the ghost neuron is connected with all neurons in the next layer with some weights $\mathbf{W}_g$, and depending on these values it can successfully reduce additive correlated noise. In Ref.~\cite{Semenova2022Chaos} this technique was applied to a simplified untrained ANN without softmax normalization, and final performance was analysed in terms of SNR and not as a change in classification accuracy.

We will consider three variants how the ghost neuron can be introduced into the noisy hidden layer depending on its weights $\mathbf{W}_g$. The first way is to simply set the weights to -1, i.e. $W_{g,i} = -1$. The results of this ad-hoc technique corresponds to black dashed lines marked as ``ghost neuron I'' in Fig.~\ref{fig:ghost}(b,c). It is clear that including ghost neurons with such an un-optimized connection strength does not result in any noise mitigation. This can be explained by the fact that the impact of noise in layer $n-1$ onto any neuron $j$ in layer $n$ depends on its particular connections to the previous layer obtained during training, and therefore subtracting the noise with a constant weighting of $-1$ does not result in any particular noise reduction. 

Let us now consider the signal coming from layer $n$ to the next layer. Each $j$th neuron of layer $(n+1)$ receives the incoming signal $\sum\limits^k_{i=1}W^n_{i,j}\cdot Y^n_i$ from layer $n$. In the case of a ghost neuron, its signal is added according to 
\begin{equation}
	X^{n+1}_j = \sum\limits^k_{i=1}W^n_{i,j}\cdot Y^n_i + W_{g,j}\cdot Y_g.
\end{equation}
This can be rewritten for additive correlated noise as follows:
\begin{equation}
	\begin{array}{c}
		X^{n+1}_j = \sum\limits^k_{i=1}W^n_{i,j}\cdot f(X^n_i) + \sqrt{2D^C_A}\xi^C_A(t)\cdot\sum\limits^k_{i=1}W^n_{i,j} + \\ W_{g,j} f(0) + W_{g,j}\cdot\sqrt{2D^C_A}\xi^C_A(t).
	\end{array}
\end{equation}
Thus, setting
\begin{equation}\label{eq:Wg}
	W_{g,j}=-\sum\limits^k_{i=1}W^n_{i,j}
\end{equation}
will allow the complete suppression of additive additive correlated noise. This case corresponds to gray lines marked as ``ghost neuron II'' in Fig.~\ref{fig:ghost}(b,c). In both, training and testing data, this method leads to significant increase in accuracy for large noise intensity, while for small or even diminishing noise intensities the network performance becomes worse. Such an analytical rule provides an excellent tool for circuit designers of hardware ANNs. It allows to include the ghost neuron concept ad-hoc and design internal. It simply requires a circuit that sets the ghost neuron's weight in a separate stage, where all neurons in the preceding layer are clamped to one, as Eq.~(\ref{eq:Wg}) is the overall input strength experienced by the neuron. 

Taking a close look at the data for concept ghost neuron II (Fig.~\ref{fig:ghost}(b,c)), one can see that now the accuracy does not depend on the noise intensity, i.e. noise has been completely suppressed, but there is some constant offset that affects the accuracy. This is an interesting feature previously not identified. When we introduced the ghost neuron, we used a neuron with the same noise and activation function as the rest of the neurons in the hidden layer, but deprived it of an input signal. The activation function in hidden layer is the sigmoid function $f(x) = \frac{1}{1+e^{-x}}$, which for zero input leads to $f(0)=0.5$ as a noise-free output of the ghost neuron. In the case of considered here sigmoid activation function, any large negative bias value can lead to $f(x)\approx 0$, and the result of clamping the ghost neuron to a constant and large negative input is shown by the orange dashed line in Fig.~\ref{fig:ghost}(b,c) and labelled as ``ghost neuron III''. It is clear that for this correction the ghost neuron perfectly fulfils its intended duty, and the accuracy remains the same as for a noise-free ANN for any intensity of correlated additive noise. Of course, if the particular activation function of neurons in the hardware ANN is such that for $f(0)\approx 0$ (i.e. $\tanh(x)$ or ReLu), then the application of ghost neuron II will lead to the same effect as the ghost neuron III.

The way how all three ghost neurons can be implemented in Python code is described in details in Supplementary materials.

\section{Conclusions}\label{sec:conclu}
In our previous work on noisy hardware ANNs and concepts of noise mitigation, we focused on the impact of noise onto the hardware ANN's output signal to noise ration. However, the thresholded objective of an ANN when addressing classification tasks substantially changes what it is relevant. The same can be said for the here studied softmax normalization in the final ANN layer. Most importantly, we found that the common perturbation of output classifiers by correlated noise is less relevant than for ANNs addressing analog output function approximation.

We considered several types of noise in hidden and the output layer, separately, and found that a hardware ANN's accuracy is less susceptible to noise introduced into the output layer, especially if this noise is additive. This can be explained by the fact that for classification tasks usually the softmax function is applied in the last layer. In this, and in the context of other \emph{winner-takes-all} encoding in the output layer, it is not so much the particular noisy value but which neuron has the largest activity. Typically, for well-trained networks, the output of the correct neuron is much larger than the output of the remaining neurons. This generally reduces the impact of noise perturbations, as noise now first needs to overcome the separation between the largest and the other neurons, which introduces a strong thresholding on noise.

This is especially evident for additive noise in the output layer, which practically does not change the accuracy. Multiplicative noise in the output layer, however, has a stronger effect, since its effect is larger for the usually larger activity neurons. The largest impact on accuracy is demonstrated by the noise in the hidden layer, especially if it is additive noise. This is quite different from our previous results \cite{Semenova2019, Semenova2022NN}, where we did not consider performance accuracy, and the resulting conclusions substantially different in that noise in the hidden layer can be suppressed due to connectivity, and does not affect the noisiness of the output signal as much as noise in the output layer. Here we consider the influence of noise on the accuracy of the network, and the conclusion actually is inverted.

Finally, we propose several noise reduction techniques and describe how they can be implemented in a hardware ANN. We propose pooling to reduce uncorrelated noise, and show that it works very well for, both, additive and multiplicative noise. To reduce the additive correlated noise, we propose the ghost neuron technique. Initially, this technique was proposed by us in the work \cite{Semenova2022Chaos}, but here we show that this technique must take into account the peculiarities of the activation function in the noisy layer. The importance of the classification context is also relevant for the ghost neuron technique. Here we show that by adding some ghost neurons we can not only suppress the level of noise but also degrade the accuracy of the network. We therefore extended our analysis and were able to derive an analytical design rule allowing ghost neuron topology adjustment on a circuit design level. This is of substantial relevance for future hardware ANN design. 

In this paper, we were focused on a simple network architecture for clarity and apply pooling and ghost neuron techniques to this simple network with only one hidden layer. For this network we suggested how to suppress the noise level in hidden layer $2$ changing the connection matrices $\mathbf{W}^1$ and $\mathbf{W}^2$ connecting layers 1-2 and 2-3, respectively. At the same time, this can be applied to any deep neural network with any number of layers. If it is necessary to apply the noise reduction technique in layer $n$, then one need to do all the same operations with the connection matrix before this layer $\mathbf{W}^{n-1}$ as described for $\mathbf{W}^1$, and transform the following matrix $\mathbf{W}^n$ in the same way as $\mathbf{W}^2$ matrix.

\section*{Supplementary Material}
Our supplementary material contains a pdf-file with additional description of noise reduction methods and how they can be implemented using Python code. It starts with pooling technique and then we consider all three ghost neurons.

\begin{acknowledgments}
	This work was supported by the Russian Science Foundation (project No. 23-72-01094) %\hyperref{https://rscf.ru/project/23-72-01094/}
\end{acknowledgments}

\section*{Data Availability Statement}
The data that support the findings of this study are available from the corresponding author upon reasonable request.

\section*{References}
\bibliography{bibliography}% Produces the bibliography via BibTeX.
	
\end{document}